\documentclass[conference]{IEEEtran}
\IEEEoverridecommandlockouts

\usepackage{booktabs} 

\usepackage[english]{babel}
\usepackage{moresize}
\usepackage{amsmath}
\usepackage{algorithmic}
\usepackage{balance}
\usepackage{comment}
\usepackage{paralist}
\usepackage{bm}
\usepackage{pgfplots}
\usetikzlibrary{pgfplots.dateplot}

\usepackage{flushend}
\usepackage[english]{babel}
\usepackage[latin1]{inputenc}
\usepackage{mathrsfs}
\usepackage{graphicx}

\usepackage{amssymb}
\usepackage{amsfonts}
\usepackage{url}
\usepackage{longtable}
\usepackage{rotating}
\usepackage{multirow}
\usepackage{mathrsfs}
\usepackage{subfigure}
\usepackage{enumitem}
\usepackage[linesnumbered,algoruled,boxed,lined]{algorithm2e}
\usepackage{adjustbox}
\usepackage{hyperref}
\usepackage{pgfplots}
\usetikzlibrary{pgfplots.dateplot}
\usepackage{filecontents}
\usepackage{caption2}
\usepackage{marginnote}

\definecolor{tblue}{RGB}{31,119,180}
\definecolor{torange}{RGB}{255,127,14}
\definecolor{tgreen}{RGB}{44,160,44}
\definecolor{tred}{RGB}{214,39,40}
\definecolor{tpurple}{RGB}{148,103,189}

\newcommand{\hide}[1]{} 

\newcommand{\wrt}{\textit{w}.\textit{r}.\textit{t}}

\newcommand{\ICDERevision}{\color{black}}

\def\model{GraphAug}

\begin{document}



\title{Graph Augmentation for Recommendation

\noindent \thanks{\textbf{*Corresponding Author is Chao Huang.}}}


\author{\IEEEauthorblockN{Qianru Zhang$^{1}$, Lianghao Xia$^{1}$, Xuheng Cai$^{1}$, Siu-Ming Yiu$^1$, Chao Huang$^{1,*}$, Christian S. Jensen$^{2}$}
\IEEEauthorblockA{University of Hong Kong$^1$, Aalborg University$^2$ \\
\{qrzhang, lhxia, rickcai, smyiu, chuang\}@cs.hku.hk, {csj@cs.aau.dk}}}

\maketitle
\pagenumbering{arabic}
\setcounter{page}{1}
\begin{abstract}

Graph augmentation with contrastive learning has gained significant attention in the field of recommendation systems due to its ability to learn expressive user representations, even when labeled data is limited. However, directly applying existing GCL models to real-world recommendation environments poses challenges. There are two primary issues to address. Firstly, the lack of consideration for data noise in contrastive learning can result in noisy self-supervised signals, leading to degraded performance. Secondly, many existing GCL approaches rely on graph neural network (GNN) architectures, which can suffer from over-smoothing problems due to non-adaptive message passing. To address these challenges, we propose a principled framework called \model. This framework introduces a robust data augmentor that generates denoised self-supervised signals, enhancing recommender systems. The \model\ framework incorporates a graph information bottleneck (GIB)-regularized augmentation paradigm, which automatically distills informative self-supervision information and adaptively adjusts contrastive view generation. Through rigorous experimentation on real-world datasets, we thoroughly assessed the performance of our novel \model\ model. The outcomes consistently unveil its superiority over existing baseline methods. The source code for our model is publicly available at: \url{https://github.com/HKUDS/GraphAug}.

\end{abstract}

\section{Introduction}
\label{sec:intro}



In recent years, Graph Neural Networks (GNNs) have received considerable attention as powerful frameworks for Collaborative Filtering (CF). They excel at capturing complex relationships between users and items, going beyond simple pairwise associations \cite{gao2022survey, wang2019neural, he2020lightgcn}. However, most GNN-based methods rely heavily on supervised learning, which necessitates a large number of labeled examples to generate accurate user and item representations. To address this limitation, Self-Supervised Learning (SSL) has emerged as a valuable approach for recommender systems. SSL leverages unlabeled behavior data to create augmented data through pretext tasks, enabling the extraction of meaningful knowledge~\cite{yao2021self,ren2024sslrec}.

Graph Contrastive Learning (GCL) methods \cite{wu2021self, xia2022hypergraph, lin2022improving} are leading the way in self-supervised learning for recommendation systems. By effectively integrating contrastive data augmentation techniques with graph neural architectures, GCL has demonstrated its prowess in pushing the boundaries of recommender systems. The core principle of GCL lies in the maximization of mutual information \cite{zhu2021graph}. By comparing augmented views, GCL aims to enhance the consistency between the encoded embeddings of positive pairs. Simultaneously, GCL effectively separates the representations of negative samples, enabling better discrimination in the learning process.

The efficacy of data augmentation in user preference learning through Graph Contrastive Learning (GCL) critically hinges on the quality of the contrastive embeddings derived from augmented graph structural views \cite{tian2022learning, chen2021autodebias}. However, real-world recommendation environments are often plagued by interaction noise, stemming from the inherent diversity and randomness of user behaviors, including factors such as misclicks \cite{tian2022learning} and popularity bias \cite{chen2021autodebias}. Unfortunately, many existing GCL methods that perform augmentation with noisy information are prone to providing inaccurate self-supervised learning (SSL) signals, consequently leading to a distorted understanding of user preferences in recommenders. To illustrate further, the random perturbation of graph structures \cite{wu2021self, yu2022graph} can result in the loss of significant interaction patterns, particularly for long-tail items \cite{liu2020long, zhang2021model}. This loss of critical information severely undermines the recommendation performance. Consequently, relying on such noisy signals for data augmentation significantly hampers the recommender system's ability to effectively capture and comprehend user preferences.

GCL approaches aim to maintain the integrity of user-item connections by incorporating manually-designed information aggregators for generating representation views in embedding contrast \cite{xia2022hypergraph, lin2022improving}. These aggregators, such as hyperedge-based embedding fusion and EM-based cluster generation, help mitigate potential damage caused by noisy interactions. However, their manual design leads to a reliance on specialized SSL pretext tasks for data augmentation, making them susceptible to noise perturbations commonly encountered in real-world recommender systems. A significant challenge arises from the widely adopted instance self-discrimination scheme in contrastive learning, which unintentionally amplifies the effects of noise and leads to misguided self-supervised learning. This challenge becomes more pronounced when inaccurate embeddings are derived from unreliable interactions or non-adaptive pretext tasks. Consequently, interaction noise has a substantial impact on the performance of existing GCL models, resulting in suboptimal data augmentation outcomes.


In addition, many existing GCL approaches heavily rely on GNN architectures. However, these architectures are susceptible to oversmoothing issues caused by non-adaptive message passing. Oversmoothing occurs when information is excessively propagated and aggregated across the graph, leading to a loss of discriminative representations and negatively impacting the model's overall performance. Consequently, effectively tackling the oversmoothing problem becomes essential for improving the accuracy and performance of GCL approaches.

This paper introduces \model, a novel self-supervised learning recommender system that serves as a robust data augmentor for recommendation tasks. The proposed approach incorporates a Graph Information Bottleneck (GIB)-regularized augmentation paradigm to distill valuable knowledge and mitigate noisy information. To achieve this, the paper introduces a reparameterization mechanism guided by graph information bottleneck constraints for dynamically learning augmented graphs. This enables \model\ to perform compressed mapping, effectively extracting important information from user-item interactions to enhance the recommendation task. The GIB-regularized information preservation is achieved by optimizing the upper and lower bounds of the mutual information maximization principle. To enhance the model's robustness, the paper introduces mixhop graph contrastive augmentation. This technique equips \model\ with strong resilience to oversmoothing by relaxing the constraint of embedding smoothing during contrastive learning-based message propagation. The seamless integration of the information bottleneck and graph contrastive learning in \model\ demonstrates promising performance in enhancing the self-supervised learning recommendation paradigm.

In a nutshell, the main contributions of this work are:

\begin{itemize}[leftmargin=*]



\item \textbf{Objective}. Our study aims to tackle the constraints of existing recommender systems that utilize graph contrastive learning (GCL) in the presence of data noise. We focus on understanding the influence of noise and underscore the crucial role of denoising self-supervised learning in facilitating resilient graph augmentation for recommendation. \\\vspace{-0.12in}



\item \textbf{Methodology}. This work presents a novel recommendation paradigm that integrates self-supervised learning (SSL) with a GIB-regularized graph augmenter. This paradigm consists of three key modules: (i) Learnable graph sampling with reparameterization, allowing for dynamic graph sampling and adaptive adjustment of the graph structure to capture relevant information. (ii) GIB-enhanced augmentation, which ensures the preservation of crucial information and mitigates noise for denoised graph augmentation. (iii) Mixhop graph contrastive learning, which enhances the GIB-regularization and addresses potential oversmoothing in recommenders. \\\vspace{-0.12in}



\item \textbf{Experiments}. We performed comprehensive experiments on diverse recommendation datasets, encompassing various settings. The evaluation results unequivocally demonstrate the superiority of our proposed model over existing solutions, including recent state-of-the-art SSL-enhanced CF models. Additionally, we investigate the advantages of our \model\ in enhancing the robustness of recommenders.

\end{itemize}


\section{Preliminaries}


\subsection{GCL for Recommendation}

To address the scarcity of labeled data, Graph Contrastive Learning (GCL) has emerged as the dominant approach for enhancing graph-based collaborative filtering~\cite{wu2021self,zhu2021graph}. GCL focuses on learning augmented representations of users and items by contrasting pairs of instances derived from corrupted interaction graph structures. In the context of graph-based collaborative filtering, the user-item relation graph $\mathcal{G}$ is constructed based on interactions between $I$ users $\mathcal{U}={u}$ and $J$ items $\mathcal{V}={v}$. This graph represents users and items as nodes, with edges indicating observed interactions. To improve graph-based recommender systems using graph contrastive learning, data augmentors manipulate the interaction graph $\mathcal{G}$ to generate two contrastive views: $\mathcal{G}'$ and $\mathcal{G}''$, which serve as augmented versions of the original graph. The primary objective is to maximize the agreement between embeddings of positively sampled pairs, while simultaneously minimizing the similarity between negative instances~\cite{lin2022improving,ren2023disentangled}.

The GCL-based recommendation paradigm incorporates the InfoNCE-based contrastive learning loss $\textbf{NCE}(\cdot)$ and the collaborative graph encoding function $\textbf{GE}(\cdot)$ to formalize the approach in the following manner:
\begin{align}
    \mathop{\arg\min}_{\mathbf{\Theta}} \mathcal{L}_\text{CL} = \textbf{NCE}\left(\textbf{GE}(\textbf{Aug}'(\mathcal{G})), \textbf{GE}(\textbf{Aug}''(\mathcal{G}))\right)
\end{align}
\noindent The application of the InfoNCE contrastive loss to the user/item embeddings, encoded by the function $\textbf{GE}(\cdot)$, is observed in models such as SGL~\cite{wu2021self} and DCCF~\cite{ren2023disentangled}. The introduction of variations to the interaction graph $\mathcal{G}$ is the responsibility of the data augmentor $\textbf{Aug}(\cdot)$. The trainable model parameters are denoted as $\mathbf{\Theta}$. Despite the SSL signals offered by GCL, the development of a noise-resistant GCL paradigm that serves as a robust data augmentor for recommendation remains a challenging and unresolved task.

\subsection{Graph Information Bottleneck}
The Information Bottleneck (IB) paradigm aims to create a mapping function that compresses input data while retaining relevant information associated with the target output labels~\cite{alemideep,saxe2019information}. Recent studies~\cite{wu2020graph,yang2021heterogeneous} have applied the IB principle to graph learning to enhance node representations. Specifically, the Graph Information Bottleneck (GIB) approach focuses on maximizing the sufficiency between the learned representation $\textbf{Z}'$ and the target task labels $\textbf{Y}$, while minimizing the mutual information between $\textbf{Z}'$ and the input graph $\mathcal{G}=(\mathcal{U}\cup\mathcal{V}, \textbf{A})$. The objective is to learn a concise yet informative representation. The GIB principle is defined:
\begin{align}
    \label{eq:infor_bottle}
   \mathop{\arg\min}_{\textbf{Z}'} \mathcal{L}_\text{GIB}= -I(\textbf{Z}';\textbf{Y}) + \beta \cdot I(\textbf{Z}';\textbf{A}), ~\beta \geq 0
\end{align}
\noindent The mutual information function is denoted by $I(\cdot)$. The trade-off between sufficiency and minimality is regulated by the Lagrange multiplier $\beta$, where larger values indicate more compressed representations. Through GIB regularization, hidden representations are optimized to achieve accurate predictions while filtering out redundant noise from the input graph $\mathcal{G}$. This regularization technique enhances performance on the downstream recommendation task.

\section{Methodology}
\label{sec:solution}

\vspace{-0.02in}
\subsection{Overview of the Proposed \model\ Framework}

\begin{figure*}
    \centering
    \hspace{-5.5mm}
      \begin{minipage}{0.99\textwidth}
    	\includegraphics[width=\textwidth]{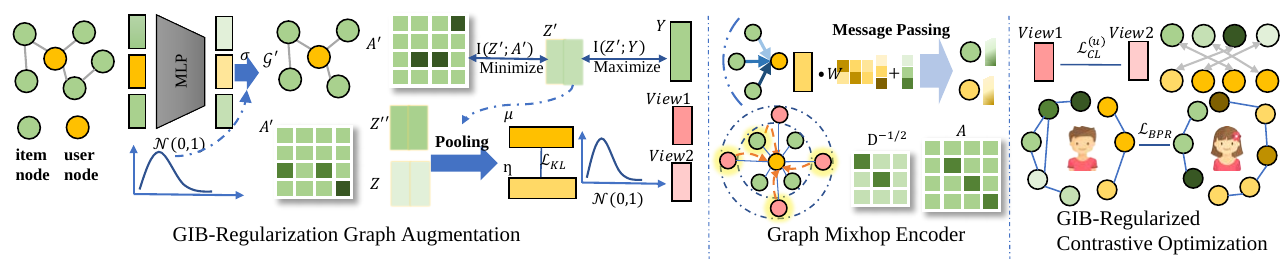}
      \end{minipage}\hspace{-2.0mm}
    \vspace*{-1mm}
    \marginnote{\color{red}}
    \caption{{\ICDERevision{The overall architecture of the \model\ framework. (i) The GIB-regularized graph augmentation is realized by integrating the learnable augmentor $\textbf{Aug}(\mathcal{G})$ and graph sampling reparameterization. (ii) Graph mixhop encoder enables mixing high-order relations for adaptive message passing. (iii) GIB-regularized contrastive optimization ($\mathcal{L}_\text{CL}$) optimizes \model.}}}
    \vspace*{-0.8mm}
\label{fig:framework}
\end{figure*}

In this section, we present a comprehensive overview of our \model, which effectively addresses two primary challenges encountered in graph-based recommendation systems: noise in the observed interaction graph and the over-smoothing problem commonly associated with existing GCL-enhanced recommender systems. The architecture of \model\ is visually depicted in Figure~\ref{fig:framework} and comprises two key components: GIB-regularization graph augmentation and mix-hop graph contrastive learning. The main objective of the first component is to denoise the observed interaction graph and enhance it by introducing additional edges that capture higher-order collaborative signals. This augmentation process aims to improve the overall quality of the graph. On the other hand, the second component of our framework focuses on learning expressive node embeddings by effectively propagating information through the augmented graph. It specifically addresses the over-smoothing issue that is often observed in GCL. \\\vspace{-0.12in}


\noindent \textbf{GIB-Regularization Graph Augmentation} utilizes GIB regularization, which encourages embeddings of similar nodes to be close in the embedding space. This regularization technique is employed to denoise the observed interaction graph and generate two denoised augmented graphs, namely $\mathcal{G}'$ and $\mathcal{G}''$. These augmented graphs include additional edges that capture higher-order collaborative relationships. \\\vspace{-0.12in}


\noindent \textbf{Graph Mixhop Encoder}. The mix-hop GNN is specifically designed to learn expressive node embeddings by effectively propagating information through the augmented graph while mitigating the over-smoothing issue that can arise in vanilla GNNs. The mix-hop GNN achieves this by incorporating mixed embeddings from different hops, enabling the capture of higher-order collaborative relationships. In particular, we employ the mix-hop propagation method to encode the augmented graphs $\mathcal{G}'$ and $\mathcal{G}''$. This approach facilitates the generation of contrastive views and enables the learning of more distinct embeddings with reduced over-smoothing effects. \\\vspace{-0.12in}

\subsection{Robust Graph Augmentation}
In order to mitigate the impact of noise in the observed interaction graph $\mathcal{G}$, our proposed \model\ framework employs an adaptive graph augmentation technique that generates a denoised version of $\mathcal{G}$. This augmentation process is further enhanced by a regularization method based on the graph information bottleneck, serving as the denoising module. Our objective is to produce a denoised graph augmentation $\mathcal{G}'$ that preserves the most informative and relevant characteristics of $\mathcal{G}$. To formalize this process, we can outline our GIB-regularized graph augmentation approach as follows:
\begin{align}
    &\mathcal{G}'=\textbf{Aug}(\mathcal{G}) = \varphi\left(p(\mathcal{G}'|\textbf{GE}(\mathcal{G}))\right)\nonumber\\
&\text{s.t.}~\mathcal{G}'=\mathop{\arg\min}_{\textbf{Z}'=\textbf{GE}(\mathcal{G}')}\mathcal{L}_\text{GIB}
\label{eq:GIBregularized}
\end{align}
\noindent The augmented graph $\mathcal{G}'$ is generated through the utilization of the function $\varphi(\cdot)$, which operates based on the embedding-based probability $p$. This probability, denoted as $p(\mathcal{G}'|\textbf{GE}(\textbf{G}))$, quantifies the likelihood of sampling the graph structure $\mathcal{G}'$ given the encoded graph embeddings of $\mathcal{G}$. To effectively incorporate the principles of the graph information bottleneck into the process of denoised graph augmentation, \model\ involves minimizing the loss $\mathcal{L}_\text{GIB}$ to facilitate the denoising of the graph encoder $\textbf{GE}(\cdot)$ across the augmented graph $\mathcal{G}'$. \\\vspace{-0.12in}

\subsubsection{\bf Learnable Augmented Graph Generation}
In \model, our primary aim is to automate the process of contrastive graph augmentation for user-item interaction graphs, allowing for adaptability across various recommendation scenarios. To initiate this process, we start by initializing an id-corresponding embedding matrix $\textbf{H}^0\in\mathbb{R}^{(I+J)\times d}$, where $I$ and $J$ denote the number of users and items in the graph, respectively, and $d$ represents the embedding dimension. Each row of $\textbf{H}^0$ corresponds to the initial embedding vector for a user $u$ or an item $v$, indicated as $\textbf{h}_u^0$ and $\textbf{h}_v^0$, respectively. Subsequently, we employ a graph encoder denoted as $\textbf{GE}$ to generate high-order node representations, resulting in the matrix $\bar{\textbf{H}}=\textbf{GE}(\textbf{H}^0)$. These representations effectively capture more intricate relationships between nodes present in the graph, thereby providing a richer and more informative graph representation.

To generate the augmented graph $\mathcal{G}'$, we employ a learnable graph augmentor denoted as $\textbf{Aug}(\mathcal{G})$. It takes the original interaction graph $\mathcal{G}$ as input and generates a modified version of the graph. The design of the augmentor allows for adaptability across different recommendation scenarios, facilitating the generation of informative graph augmentations tailored to specific recommendation tasks. The probability of an augmented graph $\mathcal{G}'$ can be decomposed into three components: the probability of the original graph $\mathcal{G}$, the probability of the perturbation process, and the probability of obtaining the resulting augmented graph $\mathcal{G}'$ given the perturbed graph.
\begin{align}
    \label{eq:prob}
    &p(\mathcal{G}'|\bar{\textbf{H}}) = \prod_{(u,v)\in\mathcal{G}'} p\left((u,v)|\bar{\textbf{H}}\right) = \prod_{(u,v)\in\mathcal{G}'}\sigma\left(\text{MLP}(\tilde{\textbf{h}}_u || \tilde{\textbf{h}}_v)\right)\nonumber \\
    &\tilde{\textbf{h}}_u = (\bar{\textbf{h}}_u - \epsilon_u) \cdot \textbf{m}_u + \epsilon_u;\tilde{\textbf{h}}_v = (\bar{\textbf{h}}_v - \epsilon_v) \cdot \textbf{m}_v + \epsilon_v
\end{align}
The probability $p((u,v)|\bar{\textbf{H}})$ represents the likelihood of edge $(u,v)$ being present in the augmented graph $\mathcal{G}'$, given the user and item embeddings $\bar{\textbf{H}}\in\mathbb{R}^{(I+J)\times d}$. It is estimated using a Multi-Layer Perceptron (MLP) that takes the concatenated embeddings $\tilde{\textbf{h}}_u, \tilde{\textbf{h}}_v \in \mathbb{R}^d$ as input and produces a transformed scalar score using the sigmoid function $\sigma(\cdot)$. By adaptively incorporating noise into the original embeddings $\bar{\textbf{h}}_u, \bar{\textbf{h}}_v$ and utilizing binary mask vectors $\textbf{m}_u, \textbf{m}_v$, the learnable graph augmentor distills informative graph features, thereby improving the accuracy of probability estimation. The derived edge sampling probability $p(\mathcal{G}'|\textbf{GE}(\textbf{G}))$ effectively mitigates the impact of noisy user-item interaction patterns.

\subsubsection{\bf Graph Sampling with Reparameterization}
In order to capture the distribution of the graph reflected in the encoded node embeddings, we utilize the reparameterization mechanism to sample two graph augmentations, namely $\mathcal{G}'$ and $\mathcal{G}''$, from the same distribution $p(\mathcal{G}'|\bar{\textbf{H}})$. These samples are generated with different random disturbances. It is important to note that the sampling of different edges is independent of each other, as stated in Eq~\ref{eq:prob}. For each edge $(u,v)$, we obtain a sampled graph augmentation $\mathcal{G}'$ through the reparameterization process, which can be expressed as follows:
\begin{align}
\label{eq:sample_graph}
    &\mathcal{G}' = (\mathcal{U} \cup \mathcal{V}, \textbf{A}'),~~~~ a'_{u,v}=\begin{cases}
    \bar{a}'_{u,v} ~~~\text{if}~~ \bar{a}'_{u,v}>\xi,\\
    0~~~~~~~\text{otherwise}
    \end{cases}\nonumber\\
    &\bar{{a}}'_{u,v} 
    = \sigma\left(\frac{1}{\tau_1}\left(\log\frac{p((u,v)|\bar{\textbf{H}})}{1 - p((u,v)|\bar{\textbf{H}})} + \log \frac{\epsilon'}{1-\epsilon'}  \right)\right)
\end{align}
The augmented graph $\mathcal{G}'$ is represented by the adjacent matrix $\textbf{A}'\in\mathbb{R}^{I\times J}$. Each element $a_{u,v}'$ in $\textbf{A}'$ is obtained by applying a soft-Bernoulli distribution $\bar{a}_{u,v}'$ with a threshold $\xi$. To introduce randomness, Gaussian noise $\epsilon'\sim N(0, I)$ is utilized to generate Gumbel noise. Our reparameterization-based approach enables differentiable sampling of $\mathcal{G}'$ and $\mathcal{G}''$, facilitating gradient-based denoising of user-item interaction data through GIB-regularized representation learning. \\\vspace{-0.12in}

\subsubsection{\bf GIB-enhanced Augmentation}
We employ the loss function $\mathcal{L}_\text{GIB}$ (Eq~\ref{eq:infor_bottle}) to maximize the mutual information between the compressed embeddings of the sampled augmented graphs $\mathcal{G}'$ and $\mathcal{G}''$, as well as the labels $\textbf{Y}$. However, optimizing $\mathcal{L}_\text{Aug}$ directly is challenging due to the intractability of the mutual information term $I(\cdot)$. To overcome this, we focus on optimizing upper bounds for $I(\textbf{Z}'; \textbf{A})$ and lower bounds for $I(\textbf{Z}'; \textbf{Y})$ instead. Here, we will elaborate on the calculation for $\mathcal{G}'$ and provide details on the upper bound for $I(\textbf{Z}'; \textbf{A})$ and the lower bound for $I(\textbf{Z}'; \textbf{Y})$. \\\vspace{-0.12in}

\noindent \textbf{Lemma 1: Upper bound of $I(\mathbf{Z}';\mathbf{A})$}. 
To assess the upper bound of $I(\textbf{Z}';\textbf{A})$ for the augmented graph $\mathcal{G}'$, where the sampled adjacent matrix $\textbf{A}'$ is obtained from the original matrix $\textbf{A}$, we can represent it in the following manner:
\begin{align}
\label{eq:upper_final}
I(\mathbf{Z}';\mathbf{A}) \leq \sum \sum  p(\mathbf{A}') p(\mathbf{Z}'|\mathbf{A}') \log{\frac{p(\mathbf{Z}'|\mathbf{A}')}{r(\mathbf{Z}')}}
\end{align}
We denote the low-dimensional embeddings obtained from the sampled graph $\mathcal{G}'$ and the disturbed embeddings $\tilde{\textbf{H}}$ as $\mathbf{Z}' = \textbf{GE}(\mathcal{G}', \tilde{\textbf{H}})$, where $\textbf{GE}(\cdot)$ is the graph encoder. Additionally, $r(\textbf{Z}')$ serves as a Gaussian-based approximation for $p(\textbf{Z}')$. It is important to note that the Kullback-Leibler divergence $\text{KL}(p(\textbf{Z}'), r(\textbf{Z}'))\geq 0$ holds true, as mentioned in~\cite{alemideep}. \\\vspace{-0.12in}

\noindent \textbf{Lemma 2: Lower bound of $I(\mathbf{Z}';\mathbf{Y})$}. 
Considering the input graph $\mathcal{G} = (\mathcal{U} \cup \mathcal{V}, \mathbf{A})$, the sampled augmentation $\mathcal{G}'$ along with its corresponding sampled adjacency matrix $\textbf{A}'$, and the learned embeddings $\textbf{Z}'$, the lower bound for $I(\mathbf{Z}';\mathbf{Y})$ is:
\begin{align}
    \label{eq:part_lower}
    I(\mathbf{Z}';\mathbf{Y}) \geq \sum \sum  p(\mathbf{Y},\mathbf{Z}') \log{q(\mathbf{Y}|\mathbf{Z}')} + H(\mathbf{Y})
\end{align}
\noindent We can disregard $H(\mathbf{Y})$ as it is independent of the optimization process. To approximate $p(\mathbf{Y}|\mathbf{Z}')$, the distribution $q(\mathbf{Y}|\mathbf{Z}')$ is trained, taking advantage of the non-negativity of the KL divergence $KL(p(\mathbf{Y}|\mathbf{Z}'), q(\mathbf{Y}|\mathbf{Z}'))$.

By leveraging the earlier obtained lower bound for $I(\mathbf{Z}';\mathbf{Y})$ and upper bound for $I(\mathbf{Z}';\mathbf{A})$, we can combine them using Eq~\ref{eq:infor_bottle}. This allows us to transform the learning objective into minimizing the subsequent upper bound:
\begin{align}
    \label{eq:new_objective}
    \mathcal{L}_\text{GIB}&= -I(\mathbf{Z}';\mathbf{Y}) + \beta \cdot I(\mathbf{Z}'; \mathbf{A}') \nonumber\\
    &\leq - \sum\sum p(\mathbf{Y},\mathbf{Z}')\log{q(\mathbf{Y}|\mathbf{Z}')}
    \nonumber \\
    &+ \beta \cdot \sum \sum  p(\mathbf{A}') p(\mathbf{Z}'|\mathbf{A}') \log{\frac{p(\mathbf{Z}'|\mathbf{A}')}{r(\mathbf{Z}')}}
\end{align} 

To simplify the notation, we refer to the aforementioned upper bound as $\bar{\mathcal{L}}$. We aim to minimize $\bar{\mathcal{L}}$ by approximating it using the empirical data distribution $p(\mathbf{Y}, \mathbf{A}) = p(\mathbf{A}) p(\mathbf{Y}|\mathbf{A}) = \frac{1}{N}\sum^{N}_{n=1}\delta{\mathbf{Y}_n}(\mathbf{Y})\delta{\mathbf{A}_n}(\mathbf{A})$.
\begin{align}
    \label{eq:final_obj}
    &\mathcal{L}_\text{GIB} \leq \bar{\mathcal{L}} \approx \mathcal{L}_\text{KL} \nonumber \\ 
    = &\frac{1}{N}\sum^{N}_{n=1} \left\{-\log{q(\mathbf{Y}_n|\mathbf{Z}')}+\beta\cdot p(\mathbf{Z}'|\mathbf{A}_n) \log{\frac{p(\mathbf{Z}'|\mathbf{A}_n)}{r(\mathbf{Z}')}}\right\}
\end{align}
The distribution $p(\mathbf{Z}'|\mathbf{A}_n)$, denoted as $\mathcal{N}(\mathbf{Z}'|\mu(\mathbf{A}_n), \eta(\mathbf{A}_n))$, is a Normal distribution with a mean value of $\mu(\textbf{A}_n)$ and a standard deviation of $\eta(\textbf{A}_n).$ We compute the mean and standard deviation as follows:
\begin{align}
    \label{eq:pooling}
     ({\mu}(\mathbf{A}_n), {\eta}(\mathbf{A}_n)) = \text{Pooling}(\{\mathbf{Z}', \mathbf{Z}'', \textbf{Z}\})
\end{align}
where $\mu(\textbf{A}_n) \in \mathbb{R}^{I \times d/2}$ and $\eta(\textbf{A}_n) \in \mathbb{R}^{I \times d/2}$. $\text{Pooling}(\cdot)$ denotes mean-pooing for the embeddings of different views.

{\ICDERevision{\subsection{Graph Mixhop Encoder}}}\marginpar{\color{red}}
To tackle the issue of over-smoothing encountered in GIB-regularized Graph Contrastive Learning (GCL), we propose the Graph Mixhop encoder. The Graph Mixhop encoder is specifically designed to enhance the graph contrastive learning procedure and foster the generation of resilient representations.

To address the over-smoothing issue in vanilla Graph Neural Networks (GNNs) and capture high-order collaborative relationships, we propose the Mixhop graph neural encoder. Inspired by Mix-hop GNNs~\cite{abu2019mixhop, wu2020connecting}, our approach incorporates denoised graph augmentations and mixes embeddings from different hops. This enables the capture of high-order collaborative relationships. The embeddings are generated iteratively, multiplying and adding previous-order embeddings with corresponding adjacency matrices. Finally, a learnable weight vector is used to combine the resulting embeddings.

By adopting our graph encoder $\textbf{GE}(\cdot)$ in this manner, we are able to successfully capture high-order collaborative relationships. Importantly, this approach eliminates the requirement of pre-calculating complex high-order relations of the adjacency matrix, resulting in reduced memory costs. The $(l+1)$-order embedding for user $u$ in relation to the graph augmentation $\mathcal{G}'$ can be formulated as follows:
\begin{align}
\label{eq:mix_hop}
    \textbf{h}_u^{'(l+1)} = \mathop{\Bigm|\Bigm|}_{m\in\mathcal{M}}\delta\Bigm(\sum_{\tilde{a}_{u,v}^{'m}\neq 0} \tilde{a}_{u,v}^{'m} \textbf{W}_m^{(l)}\textbf{h}_v^{'(l)}\Bigm)
\end{align} 
\noindent where $\textbf{h}_v^{'(l)} \in \mathbb{R}^d$ are the embeddings for a neighboring item $v$ of user $u$ during the $l$-th iteration of the GNN. The transformation matrix $\textbf{W}_m^{(l)} \in \mathbb{R}^{d\times d}$ is learnable, and the LeakyReLU activation function is denoted as $\delta(\cdot)$. The $\Arrowvert$ symbol signifies vector concatenation, while $\mathcal{M}$ denotes a predefined set of hops. $\tilde{a}_{u,v}^{'m}$ corresponds to the user-item pair $(u,v)$ in the $m$-th power of the matrix $\tilde{\textbf{A}}'$, denoted as $\tilde{\textbf{A}}^{'m}$. 

The matrix $\tilde{\textbf{A}}'$ is a Laplacian-normalized adjacency matrix with a self-loop, following the approach in \cite{wang2019neural}, and it is symmetric. If $\mathcal{M} = {1}$, the mix-hop GNN reduces to a vanilla GNN. When $\mathcal{M} = {0,1,2}$, the information propagation involves combining $\tilde{\textbf{A}}^{'0}$, $\tilde{\textbf{A}}^{'1}$, and $\tilde{\textbf{A}}^{'2}$. Here, $\tilde{\textbf{A}}^{0}$ represents the identity matrix $\textbf{I}$ with a dimensionality of $(I+J)$. The mix-hop message passing is conducted similarly for item nodes. \\\vspace{-0.12in}

\noindent \textbf{High-Order Smoothing via Mixhop Propagation}.
Our \model\ stands out from the majority of existing graph contrastive learning models~\cite{wu2021self, yu2022graph} by effectively addressing the over-smoothing issue through mixhop propagation. Drawing inspiration from mixhop GNNs~\cite{abu2019mixhop, wu2020connecting}, our approach places emphasis on capturing high-order collaborative relationships. In \model, we initiate the process by extracting GNN embeddings using the graph structure encoding function $\textbf{GE}(\cdot)$ from denoised graph augmentations $\mathcal{G}'$ and $\mathcal{G}''$.

In order to enhance our graph encoder and effectively address the over-smoothing issue, we introduce a mix-hop propagation method that combines embeddings from different hops. This approach enables us to capture high-order interactive relations while alleviating the over-smoothing problem. Here is a detailed overview of how the mix-hop propagation method operates within our \model: Given a user $u$ and the graph augmentation $\mathcal{G}'$, the $(l+1)$-order embedding is computed as a weighted mixture of the $l$-order embeddings for all users $v$ in the graph. The weights of the mixture are determined by the $l$-th row of a mixing matrix $\textbf{M}$. During the training process, the matrix $\textbf{M}$ is learned to optimize the downstream task and controls the contribution of different hop embeddings to the $(l+1)$-order embedding. By incorporating mix-hop propagation, our \model\ effectively tackles the over-smoothing issue, ensuring the generation of meaningful and informative embeddings. The following derivations provide a step-by-step illustration of this process.
\begin{align}
\label{eq:oversmooth_de}
\mathbf{H}^{(1)} &= \mathop{\Arrowvert}\limits_{m \in \left\{0,1,2\right\}} \sigma(\tilde{\mathbf{A}}^{m} \cdot \mathbf{X} \cdot \mathbf{W}_{j}^{(0)})\nonumber\\
&= \sigma(\mathop{\Arrowvert}\limits_{m \in \left\{0,1,2\right\}}\tilde{\mathbf{A}}^{m} \cdot \mathbf{X} \cdot \mathbf{W}_{j}^{(0)}) \nonumber\\
&= \sigma([\mathbf{I}_{N} \cdot \mathbf{X} \cdot \mathbf{W}_{0}^{(0)}|\tilde{\mathbf{A}} \cdot \mathbf{X} \cdot \mathbf{W}_{1}^{(0)}|\tilde{\mathbf{A}}^{2} \cdot \mathbf{X}  \mathbf{W}_{2}^{(0)}])
\end{align}
By utilizing the simplification where $\mathbf{W}^{(0)}_{0}$ is a zero matrix, we can further simplify the expression to $\mathbf{H}^{(1)} = \sigma([0|\tilde{\mathbf{A}}\mathbf{X}|\tilde{\mathbf{A}}^{2}\mathbf{X}])$. Building upon this, the calculation for $\mathbf{H}^{(2)}$ can be expressed as follows:
\begin{align}
\label{eq:oversmooth_de2}
\mathbf{H}^{(2)} = \sigma([\mathbf{I}_{N}\mathbf{H}^{(1)}\mathbf{W}_{0}^{(0)}|\tilde{\mathbf{A}}\mathbf{H}^{(1)}\mathbf{W}_{1}^{(0)}|\tilde{\mathbf{A}}^{2}\mathbf{H}^{(1)}\mathbf{W}_{2}^{(0)}])
\end{align}

In \model, the mixhop graph encoder plays a crucial role in generating augmented graphs $\mathcal{G}'$ and $\mathcal{G}''$. This encoder ensures the preservation of both local and global collaborative relationships during the augmentation process. By incorporating mixhop propagation, \model\ aims to effectively mitigate the over-smoothing effects commonly encountered in graph contrastive learning. This aspect sets it apart from the majority of existing graph contrastive learning models, including those referenced in~\cite{wu2021self, yu2022graph}, that do not explicitly address the over-smoothing issue through mixhop propagation. The utilization of mixhop propagation in the \model\ facilitates the learning of more distinct embeddings while minimizing the adverse impacts of graph over-smoothing. \\\vspace{-0.12in}

\subsection{GIB-Regularized Contrastive Optimization}

\marginpar{\color{red}}{\ICDERevision{\subsubsection{\bf Graph Contrastive Augmentation}

After obtaining the encoded embeddings $\hat{\textbf{H}}' = \textbf{H}^{'(L)}$ and $\hat{\textbf{H}}'' = \textbf{H}^{''(L)}$ for the augmented graphs, the subsequent step in the \model\ involves denoised graph contrastive learning. This process is achieved by minimizing a training objective that consists of two crucial components: a positive term and a negative term. The objective aims to encourage similarity between the embeddings of the same user or item from the two augmented graphs, while discouraging similarity between the embeddings of different users or items. Specifically, the cosine similarity between the embeddings of user $u$ or item $i$ from the two augmented graphs is computed as $\text{sim}(\hat{\textbf{h}}'_u, \hat{\textbf{h}}''_u)$ or $\text{sim}(\hat{\textbf{h}}'_i, \hat{\textbf{h}}''_i)$, respectively. The positive term is defined as the cumulative sum of the cosine similarities between the embeddings of the same user or item in the augmented graphs.


In contrast, the negative term is computed by randomly selecting negative samples from the sets of users and items. It involves calculating the cosine similarity between the embeddings of these chosen negative samples from the two augmented graphs. The final training objective is the summation of the positive and negative terms, with the negative term weighted by a negative sample ratio denoted as $r$. This comprehensive objective is optimized using stochastic gradient descent. By minimizing this training objective, the \model\ effectively learns and generates embeddings that possess enhanced discriminative power and resilience against graph noise and over-smoothing. Consequently, the performance of downstream tasks is significantly improved.

\begin{align}
\label{Eq:loss_cl}
    \mathcal{L}_\text{CL}^{(u)} 
    = \sum_{u\in\mathcal{U}}-\log \frac{\exp(\cos(\hat{\textbf{h}}'_u, \hat{\textbf{h}}''_u) / \tau)}{\sum_{u'\in\mathcal{U}} \exp(\cos(\hat{\textbf{h}}_u', \hat{\textbf{h}}_{u'}'') /\tau)}
\end{align}
\noindent 

In this context, the embedding vectors in $\hat{\textbf{H}}'$ and $\hat{\textbf{H}}''$ are denoted as $\hat{\textbf{h}}'$ and $\hat{\textbf{h}}''$ respectively, both belonging to the vector space $\mathbb{R}^d$. The temperature coefficient hyperparameter is represented by $\tau\in\mathbb{R}$. Specifically, $\mathcal{L}\text{CL}^{(u)}$ denotes the contrastive loss for user nodes, while a similar calculation is performed for item nodes, referred to as $\mathcal{L}\text{CL}^{(v)}$. The overall contrastive loss is determined by the sum of $\mathcal{L}\text{CL}^{(u)}$ and $\mathcal{L}\text{CL}^{(v)}$, denoted as $\mathcal{L}_\text{CL}$. By integrating our adaptive and noise-resistant augmentor, which aligns with the principles of the \model, we effectively mitigate the influence of interaction noise and task-irrelevant information, thereby preventing potential biases in the contrastive recommender system.

During the forecasting phase of our model, \model, predictions are made using user/item embeddings generated by our mix-hop graph encoder. The graph encoder is applied to the interaction graph, denoted as $\mathcal{G}$, to obtain node embeddings. The user preference score, denoted as $\hat{y}_{u,v}$, is estimated by taking the dot product of the embeddings $\hat{\textbf{h}}_u$ and $\hat{\textbf{h}}_v$, which are obtained through $\hat{\textbf{H}} = \textbf{GE}(\mathcal{G})$. For training \model, we adopt a widely-adopted pair-wise training schema~\cite{he2020lightgcn}, where triplets $(u,v^+, v^-)$ are sampled for each training sample. Here, $v^+$ and $v^-$ represent the interacted and non-interacted items of user $u$, respectively. It is ensured that $y_{u,v^+}=1$ and $y_{u,v^-}=0$, where $y$ is an element from the label matrix $\textbf{Y}$, which is identical to the interaction matrix $\textbf{A}$ in our recommendation scenario. To optimize \model, we utilize the Bayesian Personalized Ranking (BPR) loss function, which is applied to the identified training samples.
\begin{align}
\label{eq:bpr}
    \mathcal{L}_\text{BPR}=\sum_{(u,v^+, v^-)} -\log \sigma(\hat{y}_{u,v^+} - \hat{y}_{u,v^-})
\end{align}

\noindent By combining the contrastive augmentation ($\mathcal{L}\text{CL}$) with GIB-based regularization ($\mathcal{L}_\text{GIB} \approx \mathcal{L}_\text{KL}$), we establish the overall training objective for optimizing our \model\ as follows:
\begin{align}
\label{eq:joint}
    \mathcal{L} = \mathcal{L}_\text{BPR} + \beta_1\cdot \mathcal{L}_\text{GIB} + \beta_2\cdot\mathcal{L}_\text{CL} + \beta_3\cdot\|\mathbf{\Theta}\|_\text{F}^2
\end{align}
\noindent The different loss terms in the equation are weighted by $\beta_1$, $\beta_2$, and $\beta_3$. The last term in the equation corresponds to weight-decay regularization, computed as the Frobenius norm of the model parameters $\mathbf{\Theta}$. The Lagrange multiplier $\beta_1$ is selected from the range of $[1e^{-6}, 1e^{-5}, 1e^{-4}, 1e^{-3}]$. Alg 1 provides the summary of the learning steps in \model.

\begin{algorithm}[t]
	\label{alg:gibalgor}
 \caption{The \model\ Learning Algorithm}
		\KwIn{
		User-item interaction graph $\mathcal{G}$, $\tau,\tau_1$, $\beta_1, \beta_2, \beta_3$, $\xi$, learning rate $\iota$ and maximum training epochs $E$;}
		\KwOut{
		Trained node embeddings;}
		Initialize all parameters;\\
		\For{$epoch = 1, 2,..., E$}{
        Obtain high-order embedding $\bar{\mathbf{H}}$ of the input graph $\mathcal{G}$ via graph mixhop encoder;\\
        Sample graphs from $\mathcal{G}$ via reparameterization 
        following Eq~\ref{eq:sample_graph} and $\mathcal{G}', \mathcal{G}''$ are obtained;\\
        Obtain sampled graph embedding matrices $\mathbf{Z}', \mathbf{Z}''$ via Eq~\ref{eq:mix_hop} of $\mathcal{G}'$ and $\mathcal{G}''$;\\
        Learn distributions of $\mu(\mathbf{A}_n), \eta(\mathbf{A}_n)$ via pooling operation on $Z, Z', Z''$ via Eq~\ref{eq:pooling};\\
        Optimize GIB via loss $\mathcal{L}_{\text{KL}}$ according to Eq~\ref{eq:final_obj};\\
        Perform graph contrastive augmentation based on two views via loss $\mathcal{L}_{\text{CL}}^{(u)}$ according to Eq~\ref{Eq:loss_cl};\\
        Optimize BPR loss $\mathcal{L}_{\text{BPR}}$ via GIB regularizer via Eq~\ref{eq:bpr};\\
        Joint optimization of \model\ following Eq~\ref{eq:joint};\\
        }
\textbf{Return} all parameters and user \& item embeddings.
\end{algorithm}

\subsubsection{\bf Complexity Analysis of \model}

In our \model, the mix-hop iterative message passing exhibits a time complexity of $\mathcal{O}(\mathcal{M}\text{max}\times (|\tilde{\textbf{A}}'| + |\tilde{\textbf{A}}|)\times d)$ during each graph iteration. Here, $\mathcal{M}_\text{max}$ represents the maximum value of $m$ within $\mathcal{M}$, while $|\tilde{\textbf{A}}|$ and $|\tilde{\textbf{A}}'|$ denote the number of non-zero elements in $\tilde{\textbf{A}}$ and $\tilde{\textbf{A}}'$, respectively. Given that $|\tilde{\textbf{A}}'| < |\tilde{\textbf{A}}|$, the time complexity of a single mix-hop iteration can be expressed as $\mathcal{O}(\mathcal{M}_\text{max}\times |\tilde{\textbf{A}}|\times d)$. Considering $L$ iterations, the overall time complexity of mix-hop message passing becomes $\mathcal{O}(\mathcal{M}_\text{max} \times L \times |{\tilde{\textbf{A}}}| \times d)$. In real-world recommender systems, where the number of interaction records is typically large, we observe that $\mathcal{M}_\text{max}, L << |\tilde{\textbf{A}}|$. As a result, the time efficiency of our developed graph mix-hop encoder is comparable to that of vanilla Graph Neural Networks (GNNs).

Our approach offers a notable advantage by eliminating the need for pre-calculating heavy high-order relations within the adjacency matrix. Instead, we can efficiently compute $\tilde{\textbf{A}}^m\textbf{H}$ as $\tilde{\textbf{A}}(\tilde{\textbf{A}}(...(\tilde{\textbf{A}}\textbf{H})))$, resulting in reduced memory requirements. In terms of the contrastive learning loss, the computation complexity for handling negative samples is $\mathcal{O}(B\times (I+J)\times d)$. This complexity aligns with that of other contrastive learning-based recommenders~\cite{wu2021self, lin2022improving, yu2022graph}.

\section{Evaluation}
\label{sec:eval}

To comprehensively evaluate the effectiveness of our model, we conducted a series of extensive experiments, comparing it with a wide range of state-of-the-art models. Our objective was to address the following research questions:
\begin{itemize}[leftmargin=*]
\item \textbf{RQ1.} How does the performance of our \model\ compare to that of various state-of-the-art baseline methods? \\\vspace{-0.12in}
\item \textbf{RQ2.} What impact do different data views and contrastive learning components have on the prediction performance? \\\vspace{-0.12in}
\item \textbf{RQ3}: To what degree does our proposed model demonstrate resilience in the face of noise and data sparsity? \\\vspace{-0.12in}
\item \textbf{RQ4}: How does the efficiency of \model\ compare to other recommendation baselines? \\\vspace{-0.12in}
\item \textbf{RQ5}: How do different hyperparameter settings influence the performance of our \model\ framework?
\end{itemize}

\subsection{Experimental Setup}
\subsubsection{\bf Evaluation Datasets} Our model's performance was evaluated using three real-world benchmark datasets: Gowalla, Retail Rocket, and Amazon. These datasets provide valuable insights into user behavior and preferences in different domains. Here is a brief overview of each dataset: \textbf{Gowalla}: This dataset consists of users' check-in records at various locations, collected from the Gowalla platform between 2016 and 2019. It offers valuable information about user activities and preferences regarding visited places. \textbf{Retail Rocket}: The Retail Rocket dataset captures users' behavioral data, including clicks, cart additions, and transactions. It covers a 4.5-month period, providing a comprehensive view of user interactions with e-commerce platforms. \textbf{Amazon}: The Amazon dataset focuses on user ratings of products on the Amazon platform. It enables the evaluation of our model's ability to predict and recommend products based on user preferences and ratings. \\\vspace{-0.12in}

\begin{table}[t]
\renewcommand\arraystretch{1.0}
\caption{Experimental Data Statistics.}
\centering
\small
\setlength{\tabcolsep}{2.5mm}
\begin{tabular}{ccccc}
\hline
Dataset & User \# & Item \# & Interaction \# & Density \\ \hline
Gowalla &50,821      &57,440      &1,172,425             &4.0 $\times  10^{-4}$         \\
Retail Rocket     &49,611      &20,994      &169,909             &1.6 $\times  10^{-4}$         \\
Amazon &56,027      &29,525      &256,036             &1.5 $\times  10^{-4}$          \\
\hline
\end{tabular}
\label{tab:overall_data}
\vspace{-0.1in}
\end{table}

\begin{table*}[!tb]
\renewcommand\arraystretch{1.0}
  \centering
  \caption{Recommendation performance of all compared methods on different datasets in terms of Recall@20/40, NDCG@20/40.}
  \resizebox{\textwidth}{!}{
    \begin{tabular}{c|cccc|cccc|cccc}
      \toprule
            Data & \multicolumn{4}{c|}{Gowalla} & \multicolumn{4}{c|}{Retail Rocket} & \multicolumn{4}{c}{Amazon} \\
    \hline
            Metrics & Recall@20 & Recall@40 & NDCG@20 & NDCG@40 & Recall@20 & Recall@40 & NDCG@20 & NDCG@40 & Recall@20 & Recall@40 & NDCG@20 & NDCG@40\\
      \hline
      NCF& 0.1247& 0.1910& 0.0659& 0.0832&0.0812 &0.1231 &0.0456 &0.1021 &0.1032 &0.1494 &0.0453 &0.0506   \\
      AutoR& 0.1409& 0.2142& 0.0716& 0.0905&0.1019 &0.1423 &0.0578 &0.0682 &0.1213 &0.1604 &0.0605 &0.0746  \\
      \hline
      GCMC& 0.1632& 0.2346& 0.0949& 0.1140&0.1158 &0.1559 &0.0638 &0.0739 &0.1646 &0.2118 &0.1008 &0.1131   \\
      PinSage& 0.1016& 0.1296& 0.0873& 0.0971&0.0815 &0.1143 &0.0687 &0.0746 &0.0917 &0.1245 &0.0726 &0.0819  \\
      NGCF& 0.1413& 0.2072& 0.0813& 0.0987&0.1046 &0.1439 &0.0603 &0.0704 &0.1223 &0.1654 &0.0666 &0.0775  \\
      LightGCN& 0.1799& 0.2577& 0.1053& 0.1255&0.1279 &0.1642 &0.0708 &0.0824 &0.1775 &0.2276 &0.1087 &0.1216  \\
      GCCF& 0.1512& 0.2416& 0.0893& 0.1013&0.1221 &0.1656 &0.0668 &0.0778 &0.1508 &0.1994 &0.0893 &0.1018  \\
      \hline
      DisenGCN& 0.1379& 0.2003& 0.0798& 0.0961&0.1003  &0.1276  &0.0572  &0.0898  &0.1206 &0.1476 &0.0784 &0.0891 \\
      DGCF& 0.1784 & 0.2515 & 0.1069 & 0.1259 &0.1104  &0.1345  &0.0513  &0.0576  &0.1567  &0.1893  &0.1014  &0.1067   \\
      \hline
      MHCN& 0.1143& 0.1765& 0.0978& 0.1012&0.0724 &0.0961 &0.0421 &0.0469 &0.1015 &0.1345 &0.0689 &0.0785 \\
      STGCN& 0.1250& 0.1836& 0.0744& 0.0903&0.0741 &0.1120 &0.0382 &0.0479 &0.1158 &0.1536 &0.0643 &0.0849  \\
      \hline 
      SLRec& 0.1529& 0.2200& 0.0926& 0.1102&0.1268 &0.1641 &0.0732 &0.0826 &0.1679 &0.2115 &0.1045 &0.1159  \\
      SGL& 0.1814& 0.2589& 0.1065& 0.1267&0.1376 &0.1695 &0.0785 &0.0910 &0.1856 &0.2376 &0.1033 &0.1167  \\
      DGCL& 0.1793 & 0.2483 & 0.1067 & 0.1247 &0.1243  &0.1789  &0.0781  &0.0817  &0.1675  &0.2176  &0.0986  &0.1132  \\
      HCCF& 0.1818& 0.2601& 0.1061& 0.1265&0.1309 &0.1756 &0.0576 &0.0687 &0.1803 &0.2086 &0.0987 &0.1134  \\
      CGI&0.1820 &0.2514 &0.0913 &0.1120 &0.1315 &0.1723 &0.0615 &0.0712 &0.1819 &0.2214 &0.1039 &0.1205  \\
      NCL&0.1865 &0.2668 &0.1111 &0.1311 &0.1382 &0.1761 &0.0813 &0.0920 &0.1907 &0.2419 &0.1081 &0.1217  \\
      \hline
      {\model} & \textbf{0.2025} & \textbf{0.2817} & \textbf{0.1228} & \textbf{0.1442} & \textbf{0.1463} & \textbf{0.1841} & \textbf{0.0860} & \textbf{0.0956} & \textbf{0.2019} & \textbf{0.2506} & \textbf{0.1287} & \textbf{0.1414} \\
      \hline
      p-val.  & $1.3e^{-6}$ & $2.0e^{-6}$ & $2.7e^{-8}$ & $1.6e^{-8}$ & $1.4e^{-4}$ & $1.2e^{-2}$ & $1.1e^{-6}$ & $7.9e^{-6}$& $6.2e^{-5}$ & $2.0e^{-4}$ & $4.2e^{-7}$ & $3.1e^{-6}$   \\
      \hline
      \end{tabular}}
\label{tab:overall comparison}
\vspace{-0.2in}
\end{table*}

\subsubsection{\bf Baselines}
We conduct a comparative analysis of our \model\ against a range of competitive baselines that employ various recommendation paradigms.

\begin{itemize}[leftmargin=*]
\item (i) \textbf{Conventional Collaborative Filtering Approaches}. During our evaluation, we conducted a thorough comparison between our model and several representative traditional collaborative filtering (CF) methods used as baselines. The following CF methods were included in our analysis: \textbf{BiasMF}~\cite{koren2009matrix}: This widely used baseline incorporates matrix factorization to address user and item biases during the recommendation process. \textbf{NCF}~\cite{he2017neural}: NCF is a neural CF method that captures non-linear feature relationships through multiple hidden layers, enabling more expressive recommendation models. \textbf{AutoR}~\cite{sedhain2015autorec}: AutoR utilizes autoencoder-based matrix reconstruction to map user and item representations for collaborative filtering. \\\vspace{-0.12in}


\item (ii) \textbf{GNN-based Recommenders Systems}. When comparing our model to various recommendation models that utilize graph neural networks (GNNs), we consider a range of approaches. These include pioneering investigations such as GC-MC \cite{berg2017graph}, PinSage \cite{ying2018graph}, and NGCF \cite{wang2019neural}, which explore the use of graph convolutional networks to encode user-item relationships. Moreover, we examine subsequent studies that seek to simplify the GCN-based CF framework by disabling heavy non-linear transformations during the embedding propagation. Examples of such studies include LightGCN \cite{he2020lightgcn} and GCCF \cite{chen2020revisiting}. By evaluating our model against these GNN-based methods, we aim to assess its performance and determine its advantages in recommendation. \\\vspace{-0.12in}

\item (iii) \textbf{Disentangled Graph-based Recommenders}. There has been a focus on improving recommendation performance by learning fine-grained user representations and addressing intention disentanglement. We consider the following disentangled graph-based recommenders: \textbf{DGCF} \cite{wang2020disentangled}: DGCF combines user intentions to enhance recommendation performance by disentangling multiple latent factors. This approach aims to capture the diverse interests and preferences of users, leading to more personalized and accurate recommendations. \textbf{DisenGCN} \cite{ma2019disentangled}: DisenGCN utilizes a neighborhood routing mechanism to dynamically identify latent factors and learn disentangled node representations. \\\vspace{-0.12in}

\item (iv) \textbf{Generative SSL Recommendation Models}. We consider the following generative SSL methods for comparison: MHCN~\cite{yu2021self} and STGCN~\cite{zhang2019star}. These methods aim to enhance the target recommendation task by incorporating auxiliary prediction tasks. MHCN leverages hypergraph representation learning to capture global relationships, thereby augmenting SSL. On the other hand, STGCN utilizes an autoencoder to incorporate an augmented learning task focused on latent embedding reconstruction for self-supervision. \\\vspace{-0.12in}

\item (v) \textbf{Contrastive SSL Recommendation Models}. We include several recently developed contrastive SSL recommendation models in our evaluation. These models aim to enhance recommendation performance through contrastive learning. \textbf{SGL} \cite{wu2021self}: SGL applies random corruption to both the interaction graph structures and node features as a form of augmentation. \textbf{SLRec} \cite{yao2021self}: SLRec also utilizes random corruption, but focuses on node features for augmentation in contrastive SSL. \textbf{HCCF} \cite{xia2022hypergraph}: HCCF adopts local-global embedding contrasting by using the hyperedge as an information aggregator to generate global embeddings. \textbf{NCL} \cite{lin2022improving}: NCL leverages the EM algorithm to obtain global cluster representations for local-global embedding contrasting. \textbf{DGCL} \cite{li2021disentangled}: DGCL introduces disentangled contrastive augmentation with latent factor modeling. It proposes a factor-wise discriminative objective through contrastive learning.
\end{itemize}

\subsubsection{\bf Parameter settings}
The hidden dimensionality was explored within the range [8, 16, 32, 64], ensuring a comprehensive evaluation. The learning rate ($\iota$) was initialized at 0.001, accompanied by a weight decay of 0.96 to facilitate effective training. For GNN-based models, the number of message passing iterations was chosen from [1, 2, 3], allowing the models to capture high-order collaborative relationships. Additionally, the slope of LeakyReLU was fixed at 0.5 for consistent activation behavior. In terms of the InfoNCE-based contrastive loss, we experimented with different temperature parameters ($\tau$) ranging from [0.1, 0.3, 0.5, 0.7, 0.9], and the threshold ($\xi$) for graph edge sampling was carefully selected from [0.0, 0.2, 0.4, 0.6, 0.8]. When optimizing GIB-regularized models, we tuned the balance weight of the Kullback-Leibler divergence loss ($\mathcal{L}_\text{KL}$) within the range [$1e^{-6}$, $1e^{-5}$, $1e^{-4}$, $1e^{-3}$], while setting the weights ($\beta_2$ and $\beta_3$) for the contrastive learning loss and regularization term to 1.0. Lastly, we employed a weight decay regularizer of $1e^{-7}$ to prevent overfitting.

\subsection{Performance Comparison Results}
This section presents a comprehensive analysis of the performance of our \model\ model compared to different baselines across multiple datasets (refer to Table~\ref{tab:overall comparison}). Through this analysis, we draw several noteworthy observations that shed light on the effectiveness of our approach.

\begin{itemize}[leftmargin=*]
\item Our proposed method showcases better performance, surpassing all baseline models and affirming its effectiveness in recommendation tasks. While recent advancements in SSL recommender systems have incorporated generative and contrastive augmentation strategies, they often encounter challenges when confronted with noisy and sparse user-item interactions in real-world recommendation scenarios. Models like SLRec and SGL adopt stochastic feature augmentation techniques, which run the risk of information loss due to data corruption. On the other hand, contrastive approaches like NCL and HCCF rely on aligning local-level and global-level views, but the presence of interaction noise can disrupt hypergraph message passing or user clustering, thereby compromising the efficacy of SSL augmentation. These observations highlight the robustness and superiority of our proposed method over existing approaches.\\\vspace{-0.12in}


\item Moreover, the notable difference in performance between SSL-enhanced methods and GNN-based recommenders emphasizes the importance of self-supervision signals in effectively encoding user representations, especially in situations where labeled data is scarce. While building recommender systems directly on graph neural networks has its advantages, it also encounters limitations. Sparse training labels pose challenges for accurately recommending items to users with limited data, and the issue of over-smoothing leads to user embeddings that fail to capture personalized preferences. In contrast, our \model\ takes a step forward by utilizing robust augmentation techniques to address the negative impact of noisy self-supervision signals and enhances the effectiveness of the recommender system. \\\vspace{-0.12in}


\item Upon careful analysis of the results across different datasets, we observe a larger performance improvement on the ``Retail Rocket'' dataset compared to ``Gowalla'' and "Amazon." This highlights the effectiveness of our \model\ in addressing the challenges posed by extremely sparse user-item interactions, reaching a density degree as low as 1.6 $\times 10^{-4}$. In such scenarios, existing SSL methods may encounter difficulties in preserving crucial user-item interaction patterns during the augmentation process, resulting in a degradation of performance. Additionally, our graph mixhop encoder successfully recalibrates the learned augmented representations, ensuring their ability to accurately capture the uniformity of user preferences. This contribution plays a key role in enhancing the performance of our model, particularly in scenarios with highly sparse data.
\end{itemize} 

\subsection{Ablation Study}
To assess the effectiveness of our \model, we performed a comprehensive ablation study, examining the influence of modifying key components and varying augmentation strength on the model's performance. The outcomes of this study are presented in Figure~\ref{fig:ablation} and Table~\ref{tab:ab_aug}. \\\vspace{-0.12in}

\begin{figure}[t]
    \centering
    \subfigure[Ablation study of Gowalla]{
        \includegraphics[width=0.43\columnwidth]{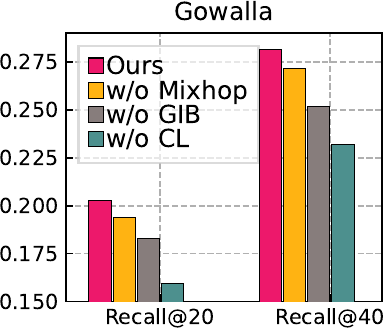}\ 
        \includegraphics[width=0.43\columnwidth]{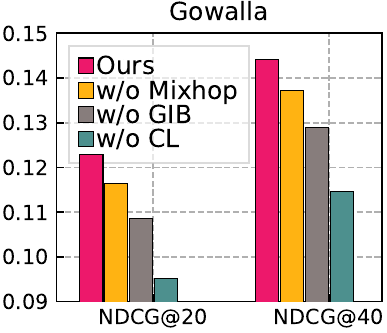}
        \vspace{-0.15in}
    }
    \subfigure[Ablation study of Retail Rocket]{
        \includegraphics[width=0.43\columnwidth]{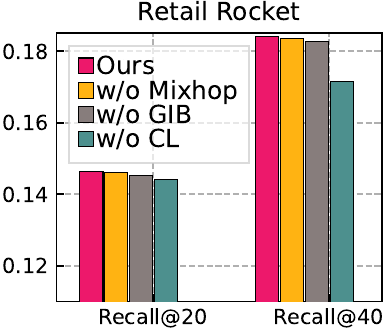}\ 
        \includegraphics[width=0.43\columnwidth]{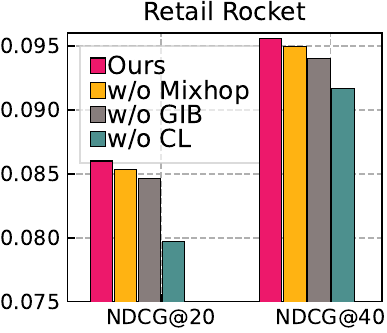}
    }
    \caption{Ablation study of sub-modules in \model.}
    \vspace{-0.15in}
    \label{fig:ablation}
\end{figure}


\begin{table}[t]
\renewcommand\arraystretch{1.0}
\centering
\setlength{\abovecaptionskip}{0.2cm}
\setlength{\belowcaptionskip}{0.1cm}
\setlength{\tabcolsep}{6.0pt}
\footnotesize
\marginnote{\color{red}}
\setcaptionwidth{0.84\columnwidth}
\caption{{\ICDERevision{Ablation study of Mixhop \wrt\ MAD}}}
\begin{tabular}{|c|ccc|}
\hline
           & \multicolumn{3}{c|}{Gowalla}                                        \\ \hline
Metrics    & \multicolumn{1}{c|}{MAD} & \multicolumn{1}{c|}{Recall@20} & NDCG@20 \\ \hline
w Mixhop   & \multicolumn{1}{c|}{{\ICDERevision{\textbf{0.7215}}}}    & \multicolumn{1}{c|}{\textbf{0.2025}}          &\textbf{0.1228}         \\ \hline
w/o Mixhop & \multicolumn{1}{c|}{{\ICDERevision{0.5783}}}    & \multicolumn{1}{c|}{0.1833}          &0.1172         \\ \hline
\end{tabular}
\label{tab:ablation_mixhop}
\vspace{-0.2in}
\end{table}

\begin{table}[t]
\renewcommand\arraystretch{1.0}
\centering
\setlength{\abovecaptionskip}{0.2cm}
\setlength{\belowcaptionskip}{0.1cm}
\setlength{\tabcolsep}{2.0pt}
\footnotesize
\marginnote{\color{red}}
\caption{Graph sampling reparameterization ablation study}
\label{tab:ab_aug}
\begin{tabular}{c|cccc}
\hline
Gowalla & \multicolumn{4}{c}{Graph Sampling Reparameterization Stregnth}                                                                  \\ \hline
Aug Ratio/Metrics           & \multicolumn{1}{c|}{Recall@20} & \multicolumn{1}{c|}{Recall@40} & \multicolumn{1}{c|}{NDCG@20} & NDCG@40 \\ \hline
0.0              & \multicolumn{1}{c|}{0.1757}          & \multicolumn{1}{c|}{0.2499}          & \multicolumn{1}{c|}{0.1051}        &0.1248         \\ \hline
0.2              & \multicolumn{1}{c|}{\ICDERevision{\textbf{0.2025}}}          & \multicolumn{1}{c|}{\ICDERevision{\textbf{0.2817}}}          & \multicolumn{1}{c|}{\ICDERevision{\textbf{0.1228}}}        &\ICDERevision{\textbf{0.1442}}         \\ \hline
0.4               & \multicolumn{1}{c|}{0.1975}          & \multicolumn{1}{c|}{0.2758}          & \multicolumn{1}{c|}{0.1198}        &0.1406         \\ \hline
0.6               & \multicolumn{1}{c|}{0.1764}          & \multicolumn{1}{c|}{0.2504}          & \multicolumn{1}{c|}{0.1059}        &0.1257         \\ \hline
0.8               & \multicolumn{1}{c|}{0.1878}          & \multicolumn{1}{c|}{0.2645}          & \multicolumn{1}{c|}{0.1128}        &0.1336         \\ \hline
Retail Rocket & \multicolumn{4}{c}{Graph Sampling Reparameterization Stregnth}                                                                       \\ \hline
Aug Ratio/Metrics           & \multicolumn{1}{c|}{Recall@20} & \multicolumn{1}{c|}{Recall@40} & \multicolumn{1}{c|}{NDCG@20} & NDCG@40 \\ \hline
0.0              & \multicolumn{1}{c|}{0.1398}          & \multicolumn{1}{c|}{0.1740}          & \multicolumn{1}{c|}{0.0827}        &0.0915         \\ \hline
0.2               & \multicolumn{1}{c|}{\ICDERevision{\textbf{0.1463}}}          & \multicolumn{1}{c|}{\ICDERevision{\textbf{0.1841}}}          & \multicolumn{1}{c|}{\ICDERevision{\textbf{0.0860}}}        &\ICDERevision{\textbf{0.0956}}         \\ \hline
0.4               & \multicolumn{1}{c|}{0.1427}          & \multicolumn{1}{c|}{0.1781}          & \multicolumn{1}{c|}{0.0847}        &0.0936         \\ \hline
0.6               & \multicolumn{1}{c|}{0.1424}          & \multicolumn{1}{c|}{0.1794}          & \multicolumn{1}{c|}{0.0836}        &0.0931         \\ \hline
0.8               & \multicolumn{1}{c|}{0.1395}          & \multicolumn{1}{c|}{0.1753}          & \multicolumn{1}{c|}{0.0822}        &0.0906         \\ \hline
Amazon & \multicolumn{4}{c}{Graph Sampling Reparameterization Stregnth}                                                                       \\ \hline
Aug Ratio/Metrics           & \multicolumn{1}{c|}{Recall@20} & \multicolumn{1}{c|}{Recall@40} & \multicolumn{1}{c|}{NDCG@20} & NDCG@40 \\ \hline
0.0              & \multicolumn{1}{c|}{0.1844}          & \multicolumn{1}{c|}{0.2294}          & \multicolumn{1}{c|}{0.1172}        &0.1290         \\ \hline
0.2               & \multicolumn{1}{c|}{\ICDERevision{\textbf{0.2019}}}          & \multicolumn{1}{c|}{\ICDERevision{\textbf{0.2506}}}          & \multicolumn{1}{c|}{\ICDERevision{\textbf{0.1287}}}        &\ICDERevision{\textbf{0.1414}}         \\ \hline
0.4               & \multicolumn{1}{c|}{0.1946}          & \multicolumn{1}{c|}{0.2418}          & \multicolumn{1}{c|}{0.1245}        &0.1368         \\ \hline
0.6               & \multicolumn{1}{c|}{0.1917}          & \multicolumn{1}{c|}{0.2370}          & \multicolumn{1}{c|}{0.1220}        &0.1339         \\ \hline
0.8               & \multicolumn{1}{c|}{0.1861}          & \multicolumn{1}{c|}{0.2302}          & \multicolumn{1}{c|}{0.1185}        &0.1301         \\ \hline
\end{tabular}
\vspace{-0.2in}
\end{table}



\noindent \textbf{Component-wise Effect}. In order to comprehensively evaluate the impacts of key components, we conducted a performance comparison by creating three variants of our model: "w/o Mixhop," where a standard GCN framework is used instead of our graph mixhop encoder; "w/o GIB," which disables the regularization for contrastive augmentation using the graph information bottleneck (GIB); and "w/o CL," where the augmentation operator with contrastive learning is removed and GIB directly regulates the optimization of the Bayesian Personalized Ranking (BPR) loss. These variants allow us to carefully analyze the individual contributions and effects of each key component in our model.

\begin{itemize}[leftmargin=*]


\item The "w/o Mixhop" variant, in which the graph mixhop encoder is excluded, demonstrates a noticeable decrease in recommendation accuracy, highlighting the positive impact of this component. The graph mixhop encoder plays a crucial role in effectively mixing high-order relationships, thereby aggregating collaborative information and mitigating the issue of over-smoothing. As a result, it leads to the generation of more uniform user representations, ultimately contributing to enhanced recommendation accuracy. \\\vspace{-0.12in}


\item A notable disparity in performance is evident between our \model\ and the "w/o GIB" variant. The integration of graph information bottleneck (GIB)-regularized augmentation in our model serves to enhance performance by functioning as a powerful data augmentor. This augmentation technique plays a crucial role in providing noise-resistant self-supervision signals, effectively countering information bias and contributing to improved performance. \\\vspace{-0.12in}


\item The significant decline in performance without the utilization of data augmentation through embedding contrasting underscores the importance of self-supervised information. By incorporating augmented CL in conjunction with the graph information bottleneck, we observe a substantial improvement in the effectiveness of SSL, particularly in addressing the challenges posed by sparse label limitations.

\end{itemize}

\noindent \textbf{Influence of Graph Augmentation Strength}: Table~\ref{tab:ab_aug} provides insights into the impact of varying graph sampling thresholds ($\xi$) using the reparameterization scheme. The results indicate that the highest performance is attained when the augmentation ratio is set to 0.2. We posit that a larger graph sampling threshold introduces more perturbations in the edges, potentially jeopardizing the preservation of crucial collaborative relationships. Conversely, a smaller $\xi$ value may still incorporate some noise as self-supervision signals for augmentation. Therefore, a balanced augmentation ratio of 0.2 proves to be effective in achieving optimal performance.

\noindent \textbf{Influence of Mixhop on Over-smoothing}. In order to assess the impact of the mixhop mechanism on mitigating the oversmoothing problem, we conducted an ablation study specifically targeting the mixhop mechanism. To evaluate the effectiveness, we employed the MAD metric (Mean Average Distance) to analyze all node embedding pairs. The results of this analysis are presented in Table~\ref{tab:ablation_mixhop}. Our findings clearly indicate that the inclusion of the mixhop mechanism in GCN leads to a higher MAD value, while its exclusion results in a lower MAD value. This compelling evidence strongly supports the effectiveness of the mixhop mechanism in successfully addressing and mitigating the issue of oversmoothing.


\begin{figure}[t]
    \centering
    \subfigure[Retail Rocket data]{
        \includegraphics[width=0.43\columnwidth]{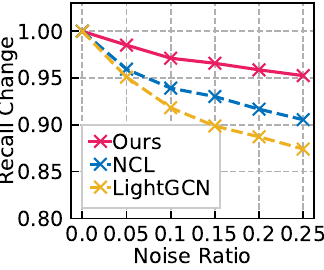}
    }
    \subfigure[Amazon data]{
        \includegraphics[width=0.43\columnwidth]{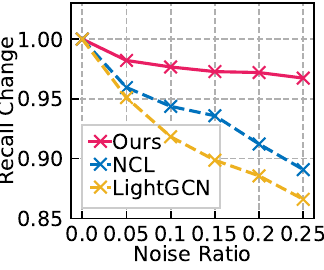}
    }
    
    \setcaptionwidth{0.84\columnwidth}
    \marginnote{\color{red}}
    \caption{\ICDERevision{The degradation in relative performance with respect to the noise ratio is examined. In this analysis, the topology of the user-item interaction graph is intentionally compromised by the introduction of randomly generated fake edges. The proportion of these fake edges is selected from the range of values: \{0.05, 0.1, 0.15, 0.2, 0.25\}.
    }}
    \vspace{-0.1in}
    \label{fig:noise}
\end{figure}

\subsection{\model-enhanced SSL Boosts Model Robustness}
The section aims to assess how well \model\ performs in the presence of structural noise and skewed data distribution.\\\vspace{-0.12in}


\noindent \textbf{(i) Against Interaction Noise}. In order to assess the performance of \model\ under the influence of structural noise and skewed data distribution, we conducted experiments. To simulate structural noise, we introduced randomly generated user-item edges to the interaction graph at varying proportions. The effect of this noise on performance is illustrated in Figure~\ref{fig:noise}, which presents the changes in Recall@20 and NDCG@20 compared to the original performance. \\\vspace{-0.12in}

\noindent \textbf{Key Findings}: Our method consistently outperforms competing approaches across various perturbation rates, exhibiting a lesser decline in performance compared to the baseline methods, namely NCL and LightGCN. This superiority can be attributed to two key factors: (1) GIB-regularized augmentation effectively aligns self-supervised learning with the recommendation task, and (2) Global relation learning incorporating mix-hop propagation mitigates the impact of noise, thereby enhancing the quality of representations and ultimately resulting in improved recommendation accuracy. \\\vspace{-0.12in}

\noindent \textbf{(ii) Against Skewed Data Distribution}.
In Table~\ref{tab:sparsity}, we present a comprehensive performance comparison between our \model\ and the baseline methods, considering varying degrees of data sparsity from both user and item perspectives. To highlight the skewed interaction distribution, we divided the training data into five user groups and five item groups based on the number of interactions. \\\vspace{-0.12in}

\noindent \textbf{Key Findings}. 
Our \model\ model is evaluated against multiple baselines, and the results, presented in Figure~\ref{fig:efficiency}, demonstrate its superior performance in terms of both convergence speed and Recall@20 and NDCG@20 metrics. This can be attributed to two key factors: the denoised SSL augmentation, which provides improved gradients for model optimization, and the information bottleneck principle, which enables robust data augmentation and effective correlation with target labels.

Additionally, we compare the performance of \model\ with other contrastive learning methods, including DGCL and NCL. Notably, DGCL exhibits the slowest convergence speed due to its disentangled representations, which pose challenges in achieving convergence with a large parameter size. On the other hand, the success of NCL heavily relies on accurate clustering results among users and items, which can be biased towards high-degree nodes, leading to misleading self-supervision information. In contrast, our proposed \model\ model achieves higher accuracy compared to the baseline methods, particularly for low-degree users and items. This highlights the effectiveness of the GIB-enhanced contrastive learning paradigm in mitigating the issue of label scarcity. The robust data augmentor incorporated in \model\ plays a crucial role in providing effective self-supervision information, thereby contributing to high accuracy in recommendation.



\begin{table}[t]
\renewcommand\arraystretch{1.0}
\centering
\setlength{\abovecaptionskip}{0.2cm}
\setlength{\belowcaptionskip}{0.1cm}
\setlength{\tabcolsep}{1.5pt}
\footnotesize
\marginnote{\color{red}}
\caption{Performance against skewed data distribution.}
\label{tab:sparsity}
\begin{tabular}{c|c|c|c|c|c|c|c}
\hline
                       &Methods                           &Metrics/Groups           & 0-10   & 10-20  & 20-30  & 30-40  & 40-50  \\ \hline
\multirow{8}{*}{Items} & \multirow{2}{*}{LightGCN} & Recall@40 & 0.0021 & 0.0103 & 0.0474 & 0.0711 & 0.0819 \\ \cline{3-8} 
                       &                           & NDCG@40   & 0.0006 & 0.0025 & 0.0117 & 0.0192 & 0.0269 \\ \cline{2-8} 
                       & \multirow{2}{*}{DGCL}     & Recall@40 & 0.0039 & 0.0172 & 0.0623 & 0.0929 & 0.0971 \\ \cline{3-8} 
                       &                           & NDCG@40   & 0.0012 & 0.0043 & 0.0159 & 0.0253 & 0.0325 \\ \cline{2-8} 
                       & \multirow{2}{*}{NCL}      & Recall@40 & 0.0450 & 0.0839 & 0.1374 & 0.1803 & 0.2170 \\ \cline{3-8} 
                       &                           & NDCG@40   & 0.0123 & 0.0230 & 0.0375 & 0.0503 & 0.0654 \\ \cline{2-8} 
                       & \multirow{2}{*}{Ours}     & Recall@40 & \ICDERevision{\textbf{0.0472}} & \ICDERevision{\textbf{0.1109}} & \ICDERevision{\textbf{0.1968}} & \ICDERevision{\textbf{0.2582}} & \ICDERevision{\textbf{0.3000}} \\ \cline{3-8} 
                       &                           & NDCG@40   & \ICDERevision{\textbf{0.0129}} & \ICDERevision{\textbf{0.0318}} & \ICDERevision{\textbf{0.0603}} & \ICDERevision{\textbf{0.0833}} & \ICDERevision{\textbf{0.1042}} \\ \hline
\multirow{8}{*}{Users} & \multirow{2}{*}{LightGCN} & Recall@40 & 0.1659 & 0.1536 & 0.1399 & 0.1206 & 0.1211 \\ \cline{3-8} 
                       &                           & NDCG@40   & 0.0702 & 0.0703 & 0.0717 & 0.0680 & 0.0662 \\ \cline{2-8} 
                       & \multirow{2}{*}{DGCL}     & Recall@40 & 0.1896 & 0.1765 & 0.1634 & 0.1444 & 0.1306 \\ \cline{3-8} 
                       &                           & NDCG@40   & 0.0812 & 0.0814 & 0.0830 & 0.0798 & 0.0769 \\ \cline{2-8} 
                       & \multirow{2}{*}{NCL}      & Recall@40 & 0.2902 & 0.2537 & 0.2255 & 0.2024 & 0.1825 \\ \cline{3-8} 
                       &                           & NDCG@40   & 0.1206 & 0.1158 & 0.1134 & 0.1094 & 0.1091 \\ \cline{2-8} 
                       & \multirow{2}{*}{Ours}     & Recall@40 & \ICDERevision{\textbf{0.3478}} & \ICDERevision{\textbf{0.2973}} & \ICDERevision{\textbf{0.2676}} & \ICDERevision{\textbf{0.2434}} & \ICDERevision{\textbf{0.2209}} \\ \cline{3-8} 
                       &                           & NDCG@40   & \ICDERevision{\textbf{0.1576}} & \ICDERevision{\textbf{0.1451}} & \ICDERevision{\textbf{0.1438}} & \ICDERevision{\textbf{0.1407}} & \ICDERevision{\textbf{0.1403}} \\ \hline
\end{tabular}
\vspace{-0.2in}
\label{tab:sparsity}
\end{table}

\subsection{Convergence Speed and Cost Time Evaluation} In this section, we thoroughly examine the convergence speed and associated cost time of our \model\ method compared to several baseline models. The results are presented in Figure~\ref{fig:efficiency} and {\ICDERevision{Table~\ref{tab:time}}}. Based on the evaluation results, we draw the following noteworthy observations and conclusions:


\begin{itemize}[leftmargin=*]


\item Our \model\ demonstrates the fastest convergence rate and surpasses other methods. This confirms the effectiveness of the denoised SSL augmentation employed by \model, which enhances the quality of gradients for model optimization. Moreover, the integration of the information bottleneck principle in \model\ facilitates robust data augmentation and effective correlation with target labels. \\\vspace{-0.12in}

\item Among the four CL-based methods, DGCL demonstrates the slowest convergence speed, highlighting the challenge of achieving convergence with a large parameter size due to its disentangled representations. Additionally, HCCF exhibits a lower convergence speed compared to \model\ and NCL. This can be attributed to the fact that the graph-hypergraph embedding contrasting in HCCF does not effectively filter out noisy users/items, impacting its convergence efficiency. \\\vspace{-0.12in}


\item Our method exhibits a comparable convergence speed to DGCL, HCCF, and NCL, as indicated in Table~\ref{tab:time}. However, it achieves the best performance among the methods. This can be attributed to its fast convergence speed.

\end{itemize}

Our \model\ model outperforms other state-of-the-art baselines in terms of Recall@20 and NDCG@20 metrics, showcasing superior performance and faster convergence speed. The comparison with other baselines provides valuable insights into the strengths and limitations of our approach, validating its effectiveness and competitiveness.

\begin{figure}[t]
    \centering
        \includegraphics[width=0.45\columnwidth]{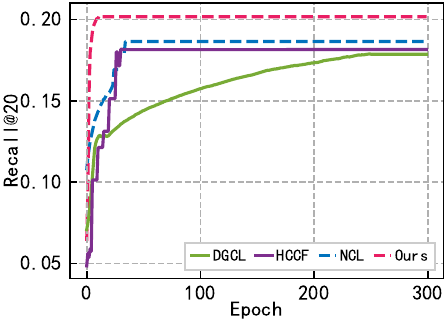}\ 
        \includegraphics[width=0.45\columnwidth]{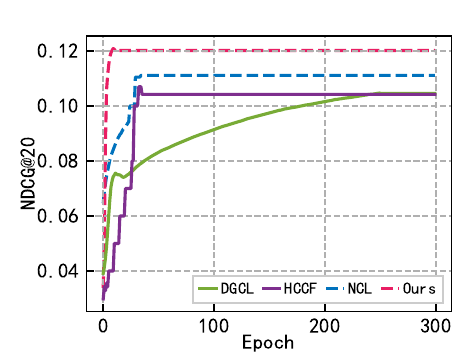}
    \caption{We examine the model convergence on the Gowalla.}
    \vspace{-0.05in}
    \label{fig:efficiency}
\end{figure}

\begin{table}[t]
\renewcommand\arraystretch{1.0}
\marginnote{\color{red}}
\setcaptionwidth{0.95\columnwidth}
\caption{{\ICDERevision{Cost time evaluation (Minutes)}}}
\label{tab:time}
\centering
\footnotesize
\setlength{\tabcolsep}{2.5mm}
\begin{tabular}{|c|cccc|}
\hline
Dataset   & \multicolumn{4}{c|}{Gowalla}                                                            \\ \hline
          & \multicolumn{1}{c|}{DGCL} & \multicolumn{1}{c|}{HCCF} & \multicolumn{1}{c|}{NCL} & Ours \\ \hline
Time      & \multicolumn{1}{c|}{{\ICDERevision{72.3058}}}     & \multicolumn{1}{c|}{{\ICDERevision{116.6662}}}     & \multicolumn{1}{c|}{{\ICDERevision{56.8012}}}    &{\ICDERevision{88.8013}}      \\ \hline
Recall@20   & \multicolumn{1}{c|}{0.1793}     & \multicolumn{1}{c|}{0.1818}     & \multicolumn{1}{c|}{0.1865}    &\textbf{0.2025}      \\ \hline
NDCG@20 & \multicolumn{1}{c|}{0.1067}     & \multicolumn{1}{c|}{0.1061}     & \multicolumn{1}{c|}{0.1111}    &\textbf{0.1228}      \\ \hline
\end{tabular}
\vspace{-0.1in}
\end{table}

\subsection{Hyperparameter Sensitivity Analysis}
In this section, we conduct a thorough investigation of the sensitivity of key parameters, including the GIB-regularized strength ($\beta$), temperature ($\tau$), and embedding dimensionality ($d$). The results of this analysis are presented in Figure~\ref{fig:hyperparameter}.

\begin{itemize}[leftmargin=*]

\item We find that the best performance is attained when setting $\beta$ to $1e^{-5}$. We utilize KL divergence as an approximation for regularization, incorporating graph information bottleneck. This approach acts as an intermediary connection, resolving the upper and lower bounds. Consequently, the recommendation accuracy may decline when the bounds for $I(Z_{\text{IB}}; Y)$ and $I(Z_{\text{IB}}; X)$ become too large or too small. \\\vspace{-0.12in}

\item Our \model\ method achieves the best performance at $\tau$ = 0.9, while performing the worst at $\tau$ = 0.5. Smaller $\tau$ values focus on hard negative pairs, while larger $\tau$ values bring similar positive samples closer. The results indicate that \model\ performs better at lower or higher temperatures. \\\vspace{-0.12in}

\item Increased hidden dimensionality enhances performance, with our proposed \model\ showing superior results and saturation at $d = 64$. However, satisfactory performance can still be attained with $d = 32$. To ensure fairness and efficiency in the comparison, we report the final results using an embedding dimensionality of 32.

\end{itemize}

\begin{figure}[t]
\centering
\begin{tabular}{c c c}
\hspace{-4.0mm}
  \begin{minipage}{0.14\textwidth}
	\includegraphics[width=\textwidth]{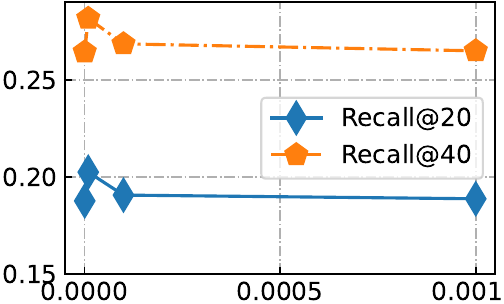}
  \end{minipage}\hspace{-2.5mm}
  &
  \begin{minipage}{0.14\textwidth}
	\includegraphics[width=\textwidth]{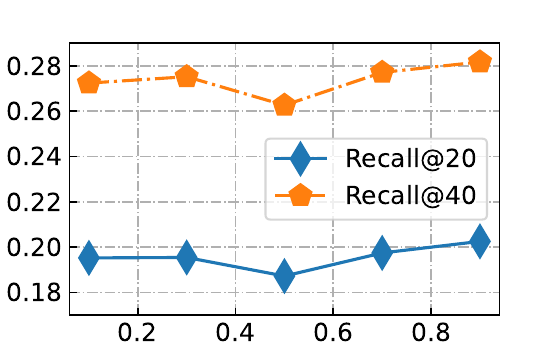}
  \end{minipage}\hspace{-2.5mm}
 &
  \begin{minipage}{0.14\textwidth}
	\includegraphics[width=\textwidth]{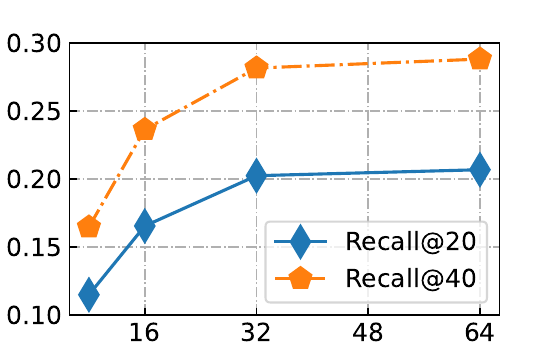}
  \end{minipage}\hspace{-2.5mm}
  \\
  (a) \# $\beta_1$-Recall
  &
  (b) \# $\tau$-Recall
  & 
  (c) \# dim-Recall
\end{tabular}
\vspace*{-0.5mm}
\caption{Hyperparameter study of \model\ on Gowalla.}
\vspace{-0.05in}
\label{fig:hyperparameter}
\end{figure}

\subsection{Case Study}
In this section, we evaluate the effectiveness of our method in denoising noisy edges between users and items through self-augmented learning. The results in Figure~\ref{fig:casestudy} address two key questions: (i) Can \model\ learn unknown implicit dependencies between items? (ii) Can \model\ identify and remove noisy edges using SSL-based augmentation?

To address these questions, we conduct an examination of the learned embeddings of three users and their interacted items from the Amazon dataset. In this analysis, each user is surrounded by their associated interacted items, and the color of each item represents a scalar value derived from its embedding vector. We make the following observations:\\\vspace{-0.1in}

\noindent \textbf{Learning implicit item dependency}.
In this analysis, items interacted by a user are grouped and represented by different colors, potentially indicating diverse user interests. These categorical relations are unknown to the model during training. However, post-training, items belonging to the same or related categories tend to exhibit close embeddings in the latent space. We establish connections between items with similar embeddings in the graph, represented by green and purple lines. While practical recommender systems may not always have access to external item knowledge, our \model\ framework has the capability to uncover implicit item relationships, thereby offering valuable insights. \\\vspace{-0.12in}

\noindent \textbf{Denoising user-item interaction bias}.
In the figure, we present the explicit similarity between user and item embeddings learned by our \model. Post-training, the model disregards connections to items with low similarity values, providing further evidence of \model's denoising effectiveness in addressing noisy user-item interaction bias. Notably, items 55021, 54805, and 66879 exhibit lower similarity to other co-interacted users, indicating that they may not accurately reflect the preferences of those users. Consequently, \model\ treats the connections to these items as noisy edges. These results affirm the efficacy of \model\ in effectively mitigating data noise and enhancing the quality of user-item interactions. \\\vspace{-0.12in}

\noindent \textbf{Distribution visualization of the learned embeddings}.
Figure~\ref{fig:cluster} visualizes representation distributions using the UMAP tool for embedding projection. Insights from the results are: LightGCN suffers from oversmoothing, while NCL and \model\ effectively address this issue through contrastive learning augmentation. Comparing NCL with our approach, \model\ learns better global uniform user representations while capturing personalized preferences. Our designed data augmentor, integrating GIB-regularized CL with the graph mixhop encoder, contributes to this effectiveness. \\\vspace{-0.12in}

\noindent \textbf{Comparison of MAD Values}. Table~\ref{tab:over_mad} presents the MAD values for node embedding pairs of our method, NCL, and LightGCN. Remarkably, our method achieves the highest MAD value, whereas LightGCN exhibits the lowest MAD value. This indicates that our method outperforms the others in effectively mitigating the issue of oversmoothing.


\begin{figure}[t]
    \centering
      \begin{minipage}{0.44\textwidth}
    	\includegraphics[width=\textwidth]{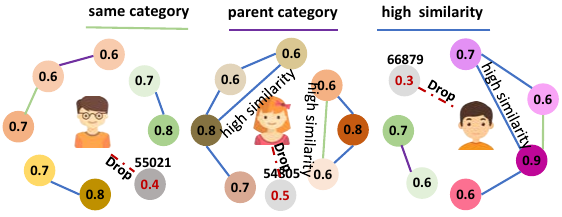}
      \end{minipage}\hspace{-3.0mm}
    \caption{In this case study, we examine the effectiveness of the proposed \model\ framework in capturing implicit item dependencies and denoising user-item interaction bias. Notably, items with similar colors indicate close embeddings, while the weights assigned to the items represent the learned similarity scores between the interacted users.}
\vspace*{-2mm}
\label{fig:casestudy}
\end{figure}

\begin{figure}[t]
\centering
\begin{tabular}{c c c }
  \begin{minipage}{0.15\textwidth}
	\includegraphics[width=\textwidth]{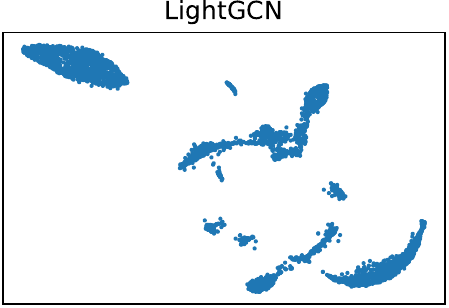}
  \end{minipage}\hspace{-3.mm}
  &
  \begin{minipage}{0.15\textwidth}
    \includegraphics[width=\textwidth]{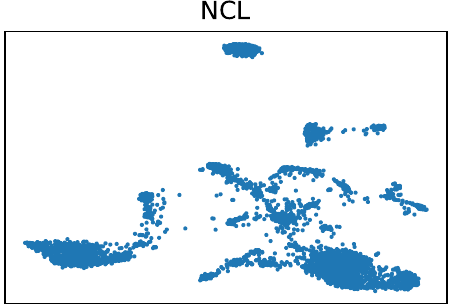}
  \end{minipage}\hspace{-3.0mm}
  &
  \begin{minipage}{0.15\textwidth}
	\includegraphics[width=\textwidth]{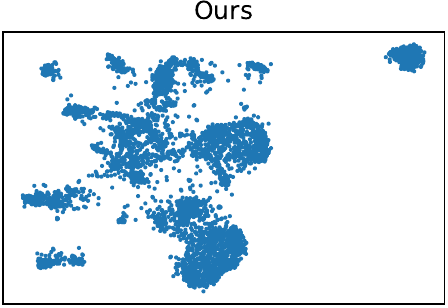}
  \end{minipage}\hspace{-3.0mm}

\end{tabular}
\caption{The visualization of embedding distributions reveals that our \model\ approach preserves better global uniformity compared to competitive baselines like LightGCN and NCL.}
\vspace{-0.1in}
\label{fig:cluster}
\end{figure}

\begin{table}[h]
\vspace{-0.05in}
\renewcommand\arraystretch{1.0}
\marginnote{\color{red}}
\setcaptionwidth{0.7\columnwidth}
\caption{{\ICDERevision{MAD value of several methods}}}
\label{tab:over_mad}
\centering
\footnotesize
\setlength{\tabcolsep}{3.0mm}
\begin{tabular}{|c|c|c|c|}
\hline
Metrics/ Methods & \model & NCL & LightGCN \\ \hline
MAD              &{\ICDERevision{\textbf{0.7215}}}      &{\ICDERevision{0.6817}}     &{\ICDERevision{0.6594}}          \\ \hline
Recall@20          &\textbf{0.2025}      &0.1865     &0.1799          \\ \hline
NDCG@20        &\textbf{0.1228}      &0.1111     &0.1053          \\ \hline
\end{tabular}
\vspace{-0.2in}
\end{table}

\section{Related Work}

\subsection{Graph Neural Networks in Collaborative Filtering}
GNNs have emerged as a leading approach in collaborative filtering, offering state-of-the-art performance~\cite{wu2022graph,gao2022survey,yang2023graph}. In general, GNN-based collaborative filtering models utilize message passing layers, where each layer is parameterized by an embedding transformation. Notably, models such as PinSage~\cite{ying2018graph} and NGCF~\cite{wang2019neural} leverage GNNs to capture high-order collaborative relations among users and items, demonstrating their effectiveness. Moreover, LightGCN~\cite{he2020lightgcn} and GCCF~\cite{chen2020revisiting} propose omitting non-linear transformations during propagation, asserting that this approach can enhance recommendation performance. Recent advancements explore alternative techniques to improve graph-based collaborative filtering, including hyperbolic representation learning~\cite{sun2021hgcf,yang2022hicf} and graph debiasing strategies~\cite{tian2022learning,fan2022graph}. Additionally, an important research direction aims to enhance the fine-grained user preference modeling by disentangling latent intent factors underlying observed user-item interactions. Prominent methods in this category include DGCF~\cite{wang2020disentangled} and DisenHAN~\cite{wang2020disenhan}.
\subsection{Self-Supervised Recommender Systems}

Existing self-supervised recommenders can be classified into contrastive and generative methods, each employing distinct strategies to improve embedding quality~\cite{xie2022contrastive,wei2021contrastive}. Contrastive learning approaches generate two contrasting views and aim to maximize the mutual information within the embeddings. For instance, SGL~\cite{wu2021self} ensures consistency between contrastive views by employing random corruption operators like node/edge dropout and random walk. Subsequent studies have further advanced contrastive learning by incorporating global information encoding through hypergraph neural networks, as demonstrated in HCCF~\cite{xia2022hypergraph}, and leveraging EM clustering algorithms, as observed in NCL~\cite{lin2022improving}. 



Generative methods harness the rich information inherent in the input data to improve embedding quality through self-supervision signals~\cite{xiacikm2021self}. One notable example is MHCN~\cite{yu2021self}, which takes inspiration from the DGI principle~\cite{velickovic2019deep}. MHCN employs a multi-channel hypergraph neural encoder to encode global-level user embeddings and aligns local and global user representations by maximizing mutual information. Another approach, AutoCF~\cite{xia2023automated}, introduces a graph masked autoencoding technique for collaborative filtering augmentation. STGCN~\cite{zhang2019star} leverages latent embedding reconstruction as a pretext task to enhance collaborative filtering. While these methods have shown effectiveness, many still rely on random or heuristic-based generation of contrastive views or predefined pretext tasks for augmentation.

\subsection{Graph Information Bottleneck}
The Graph Information Bottleneck (GIB) has gained significant attention in diverse graph representation learning applications, including graph structure learning and subgraph recognition~\cite{sun2022graph,yang2021heterogeneous,yu2022improving,yu2020graph}. For instance, Wu et al.\cite{wu2020graph} propose an extension of the general Information Bottleneck (IB) framework to improve the robustness of node and graph representations by regularizing structure or feature learning. To address graph heterogeneity, Yang et al.\cite{yang2021heterogeneous} introduce the HGIB method, which incorporates the Graph Information Bottleneck into heterogeneous graph embedding through unsupervised learning and consensus hypotheses. HGIB aims to achieve agreement between decomposed homogeneous graphs based on the mutual information maximization paradigm. To enhance the quality of graph representations, Sun et al.\cite{sun2022graph} learn node-wise connections using a tractable IB loss function. In a related work\cite{yu2022improving}, a variational GIB method is developed to model the correlations between the input graph and perturbed data. Drawing inspiration from these advancements, we propose leveraging GIB in graph contrastive learning to provide denoised data augmentation for recommender systems.

\section{Conclusion}
\label{sec:conclusion}

The paper proposes a new model called \model, which is designed to encode more robust representations using GIB-regularized contrastive learning through adaptive augmented graph generation. The model \model\ is guided by the information bottleneck principle to ensure informative self-supervision signals for the target recommendation objective. The paper demonstrates the effectiveness of the proposed model through extensive experiments, where it shows success in preserving denoised information for augmentation, which significantly boosts the recommendation performance on various datasets. Overall, the proposed model \model\ is a novel approach to encode more robust representations for recommendation systems. By utilizing GIB-regularized contrastive learning through adaptive augmented graph generation, the proposed model can capture more informative self-supervision signals that significantly improve the recommendation performance. Additionally, the future work of the paper shows the authors' dedication to exploring and broadening the model's scope in unbiased SSL and counterfactual factors in recommendation. To achieve this, the paper plans to study the complex causal correlations between the augmented interaction structures and final prediction results. This is an exciting direction for future research, as it could provide valuable insights into how the model works and how it can be improved.


\section*{Acknowledgments}
\noindent This project is partially supported by the HKU-SCF FinTech Academy. Additionally, the project is partially funded by the Shenzhen-Hong Kong-Macao Science and Technology Plan Project (Category C Project: SGDX20210823103537030) and the Theme-based Research Scheme T35-710/20-R.

\clearpage
\balance

\bibliography{sample-base}
\bibliographystyle{IEEEtran}


\end{document}